\documentclass[10pt, a4paper]{article}

\usepackage[final]{lrec-coling2024} 
\usepackage{color}

\usepackage{subfigure}
\usepackage{amsmath}
\usepackage{amssymb}

\usepackage{multirow, booktabs}
\usepackage{colortbl}
\usepackage{float}
\usepackage{tabularx}

\title{How Large Language Models Encode Context Knowledge?\\\textit{A Layer-Wise Probing Study}}

\name{Tianjie Ju\textsuperscript{1}, 
Weiwei Sun\textsuperscript{2}, 
Wei Du\textsuperscript{1}, 
Xinwei Yuan\textsuperscript{3}, \\
\textbf{\large Zhaochun Ren\textsuperscript{4}, 
Gongshen Liu\thanks{*Corresponding author.}\textsuperscript{1}*}
} 

\address{\textsuperscript{1}School of Electronic Information and Electrical Engineering, Shanghai Jiao Tong University\\
\textsuperscript{2}Shandong University, \textsuperscript{3}Southeast University, \textsuperscript{4}Leiden University\\
jometeorie@sjtu.edu.cn, sunnweiwei@gmail.com, dddddw@sjtu.edu.cn, symor@seu.edu.cn,\\ z.ren@liacs.leidenuniv.nl, lgshen@sjtu.edu.cn}

\abstract{
Previous work has showcased the intriguing capability of large language models (LLMs) in retrieving facts and processing context knowledge. However, only limited research exists on the layer-wise capability of LLMs to encode knowledge, which challenges our understanding of their internal mechanisms. In this paper, we devote the first attempt to investigate the layer-wise capability of LLMs through probing tasks. We leverage the powerful generative capability of ChatGPT to construct probing datasets, providing diverse and coherent evidence corresponding to various facts. We employ $\mathcal V$-usable information as the validation metric to better reflect the capability in encoding context knowledge across different layers. Our experiments on conflicting and newly acquired knowledge show that LLMs: (1) prefer to encode more context knowledge in the upper layers; (2) primarily encode context knowledge within knowledge-related entity tokens at lower layers while progressively expanding more knowledge within other tokens at upper layers; and (3) gradually forget the earlier context knowledge retained within the intermediate layers when provided with irrelevant evidence. Code is publicly available at \href{https://github.com/Jometeorie/probing\_llama}{https://github.com/Jometeorie/probing\_llama}.
 \\ \newline  }

\begin{document}

\maketitleabstract

\section{Introduction}
Large Language Models (LLMs), such as ChatGPT and GPT-4, have encoded massive parametric knowledge within their parameters and have achieved remarkable success in various knowledge-intensive language tasks~\citep{ChatGPT,OpenAI2023GPT4TR}. One prominent emergent capability of LLMs is their ability to encode commonsense and world knowledge acquired during the pre-training phase within their parameters, enabling them to answer factual questions directly~\citep{knowledge_retrieval_1, knowledge_retrieval_2}. 
However, the knowledge embedded during pre-training may become outdated~\citep{Nakano2021WebGPTBQ}, and the encoding of long-tail knowledge is often insufficient~\citep{Kandpal2022LargeLM, Mallen2022WhenNT}. 

Given these limitations, recent studies have focused on enhancing the factualness of LLMs using \textit{context knowledge}, such as by retrieving knowledge from the Internet or utilizing custom data~\citep{knowledge_acquisition_1, knowledge_acquisition_2}. 
Nonetheless, it is still unclear how LLMs use such context knowledge, especially when the given knowledge conflicts with their parametric knowledge.

Several studies have been dedicated to exploring the superficial capability of LLMs in utilizing context knowledge~\citep{knowledge_clashes, knowledge_clashes_2}. These studies have discovered that LLMs can alter parametric memory when exposed to coherent and non-unique evidence, while~\citet{knowledge_manipulation} doubted their capability to further utilize the context knowledge for logical reasoning. So far, there remains a notable absence of studies that delve deeply into the LLM's components to comprehensively examine the capability of intermediate layers in encoding context knowledge.

In this paper, we conduct a comprehensive investigation into the layer-wise capability of LLMs to encode context knowledge through probing tasks~\citep{probing_classifier_survey}. 
We introduce a novel dataset for this probing task, comprising coherent and diverse ChatGPT-generated evidence derived from provided facts and counterfactuals.
Subsequently, this generated evidence is fed into the LLM under explanation. Upon receiving the evidence, the outputs of its hidden layers are then extracted to serve as inputs for the probing classifier. 
The layer-wise capability of the LLM to encode context knowledge can be explained by the distinguishability of evidence from different categories within the hidden layer representations. 
We adopt $\mathcal V$-usable information~\citep{V-Usable-Information, Usable-Information} as our metric to explain the probing results, offering a more effective measure for identifying variations in context knowledge encoding across layers than test set accuracy.

Comprehensive experiments are conducted on LLaMA 2 (7B, 13B, and 70B)~\citep{LLaMA-2} to investigate the capability of LLMs in encoding context knowledge.
We initiate our study with preliminary experiments, finding that the constructed evidence does enable the LLMs to accommodate conflicting or newly acquired knowledge, especially in chat models.
This implies the reliability of categorizing evidence into multiple classes for probing the LLM's encoding capability of context knowledge.

Then, we focus on conflicting knowledge.
We draw the layer-wise heatmap of LLaMA 2 Chat 13B when dealing with the question \textit{What is Mike Flanagan's occupation?} as a case study.
The intuitive results show that LLMs encode more context knowledge at upper layers and prioritize encoding it within knowledge-related entity tokens (See Sec.~\ref{Sec: Case Study 1}).
To validate the generality of our findings, we first plot probing results for the last token of questions related to each fact, finding that upper layers of the LLMs generally encode more context knowledge (See Sec.~\ref{Sec: Average V-information}).
Then, we divide tokens in the questions into knowledge-related and non-related categories to compute the average layer-wise $\mathcal V$-information separately.
Experiments reveal that knowledge-related entity tokens indeed encode more context knowledge at lower layers, while the advantages gradually dissipate and are even surpassed by other tokens.
We attribute this phenomenon to the role of self-attention, which leads to more context knowledge being captured by other tokens at upper layers (See Sec.~\ref{Sec: Entity tokens vs. Non-Entity tokens}).

Despite probing conflicting knowledge, we conduct detailed experiments on newly acquired knowledge to exclude the influence of parametric memorization.
We design a probing task asking LLMs to answer \textit{What is the password of the president's laptop?} based on different evidence provided by contexts.
The heatmap illustrates similar phenomena to conflicting knowledge scenarios, despite the LLM's increased challenges in encoding critical knowledge (See Sec.~\ref{Sec: Case Study 2}).
Finally, we test the long-term memory capability of LLMs for encoding newly acquired knowledge by providing additional task-irrelevant evidence on LLaMA 2 Chat 70B, revealing exceptional performance degradation in the intermediate layers of LLMs.
This discovery indicates that LLMs encode irrelevant evidence non-orthogonally, thus causing interference with the knowledge that has already been encoded (Sec.~\ref{Sec: Long-Term Memory Capability}).

\begin{figure*}
  \centering
  \includegraphics[width=1.0\textwidth]{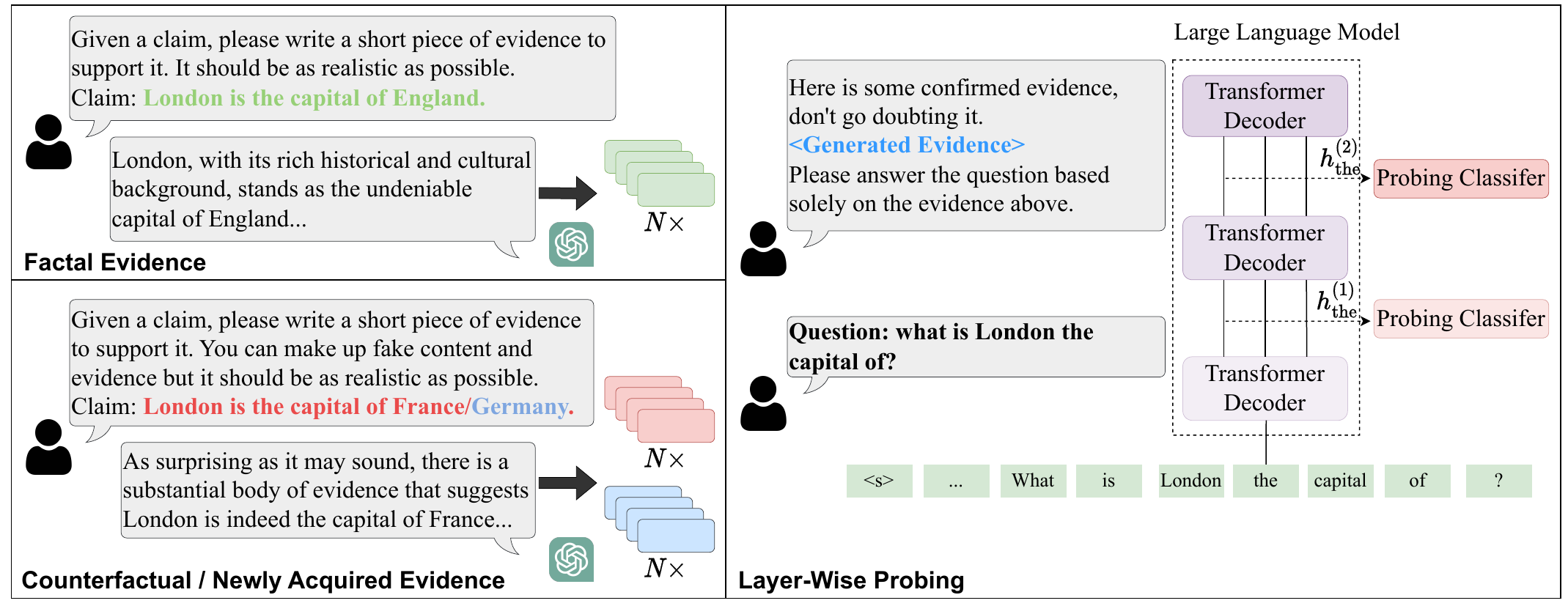}
  \caption{The overall process for probing layer-wise capability of LLMs in encoding context knowledge. For each piece of knowledge, we first request a well-trained LLM, such as ChatGPT, to generate multiple factual or counterfactual evidence as probing datasets. Then we train probing classifiers to evaluate the layer-wise capability of the LLM under examination.}
  \label{fig: methods}
\end{figure*}

\section{Related Work}
\paragraph{Explainability of LLMs}
With the broad adoption of LLMs, numerous studies have been devoted to probing their capabilities through prompting~\citep{probing_llms_1, probing_llms_2, probing_llms_3, knowledge_clashes, knowledge_clashes_2}. This line of work has exclusively attested to the exceptional capacity of LLMs in encoding context knowledge~\citep{probing_llms_4, probing_llms_5, probing_llms_6, probing_llms_7}, yet neglected to delve into the layer-wise capabilities of LLMs.

To open the black box of LLMs, \citet{edit_knowledge} proposed an approach based on causal intervention to detect memories stored in neurons. \citet{explain_gpt2_with_gpt4} devoted the first attempt to explain the neurons of GPT-2~\citep{GPT2} with the help of GPT-4, which has inspired us to the feasibility of constructing probing datasets using similar methods. Recently, \citet{probing_llm_inner} proposed RepE to monitor the high-level cognitive phenomena of LLMs, finding that LLMs tend to achieve higher neural activity when engaging in bizarre behaviors such as lying. \citet{explain_neuron} decomposed LLMs from a neuron-level perspective, discovering numerous relatively interpretable feature patterns. \citet{probing_space_and_time} first probed the capability of LLMs in encoding continuous facts, demonstrating that LLMs acquire structured knowledge such as space and time. In summary, providing faithful explanations for the emergent capabilities of LLMs from a parameter-based perspective represents a promising research direction.

\paragraph{Probing Task}
\textit{Probing task} is a promising global explanation for understanding various linguistic properties encoded in models~\citep{probing_classifier_survey, explain_llm_survey}. It usually constructs a relevant probing dataset and trains a classifier to predict certain linguistic properties from a model's representations~\citep{probing_classifier_1, probing_classifier_2, probing_classifier_3}. Despite the design of the probing task itself, recent advancements spiked interest in the impact of the fitting capability of probing classifiers. \citet{control_task} proposed control tasks by assigning random labels to inputs, thereby quantifying the performance difference between control tasks and probing tasks as selectivity, which serves as validation metrics for probing results. \citet{Information-Theoretic_Probing} approached the issue from an information theory perspective, designing control functions to calculate the difference in mutual information between the original task and the task with randomized representations. \citet{V-Usable-Information} introduced $\mathcal V$-usable information to measure dataset difficulty, which is also suitable for measuring probing datasets. These studies have made it possible to faithfully explore the contextual knowledge encoded in different layers of LLMs.

\section{Dataset Construction}
With the powerful generation capability facilitated by ChatGPT, it is no longer difficult to automatically generate diverse expressions for given context knowledge $k$. 
This capability enables us to produce an extensive array of contextual evidence $M(p \oplus k, t)$ irrespective of the truthfulness of $k$, where $p$ and $t$ represent the manually designed prompt and the temperature parameter, respectively. 
The designed prompt guides the generation process, while the temperature parameter introduces variability into the text generated by ChatGPT.

Taking the knowledge statement \textit{London is the capital of England} as an example (Fig.~\ref{fig: methods}, left), it is feasible to instruct ChatGPT to generate multiple instances of realistic factual evidence by incorporating specific prompts. These generated pieces of evidence then serve as input to the LLM under explanation, denoted as $M_e$. Furthermore, we can modify the prompts to instruct ChatGPT to provide counterfactual evidence such as \textit{London is the capital of France} or newly acquired knowledge such as \textit{The password of the president's laptop is \{password\}}.

Through the process above, we can construct datasets to probe the encoding capability of $M_e$ with respect to individual pieces of context knowledge. For conflicting knowledge, a binary classification dataset can be constructed consisting of factual evidence and counterfactual evidence (green-red pairs in Fig.~\ref{fig: methods}, left). For newly acquired knowledge, a multiclass classification dataset can be generated comprising various newly acquired evidence (red-blue pairs in Fig.~\ref{fig: methods}, left).

\section{Layer-wise Probing}
\label{sec: Layer-wise Probing}
We then provide various pieces of evidence corresponding to individual knowledge as input to $M_e$ and request it to answer the question based solely on the provided evidence (see Fig.\ref{fig: methods}, right). Since questions concerning individual knowledge remain constant, it is feasible to train a probing classifier based on the output of each hidden layer for tokens within the questions.

Specifically, for each token $w$ within the given question, we extract the output representations $R_w^{(i)}$ from the $i$-th hidden layer, which serves as input to the probing classifier $M_\textrm{probe}$. The knowledge categories $Y$ corresponding to different evidence are employed as labels for $M_\textrm{probe}$. By measuring the performance of $M_\textrm{probe}$ in learning the mapping $R_w^{(i)} \rightarrow Y$, we can infer the extent to which the hidden layer encodes context knowledge.

\begin{figure*}[t]
	\centering
	\subfigure{
	\begin{minipage}{0.31\linewidth}
    	\label{layer 1}
    	\centering
    	\includegraphics[width=1\linewidth]{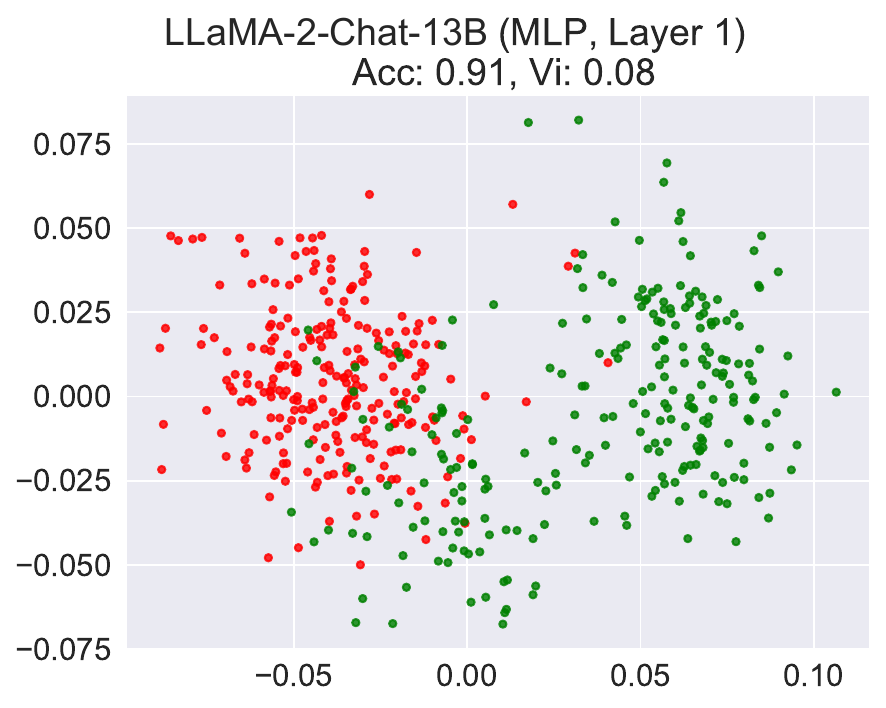}
	\end{minipage}
	}
	\subfigure{
	\begin{minipage}{0.31\linewidth}
    	\label{layer 11}
    	\centering
    	\includegraphics[width=1\linewidth]{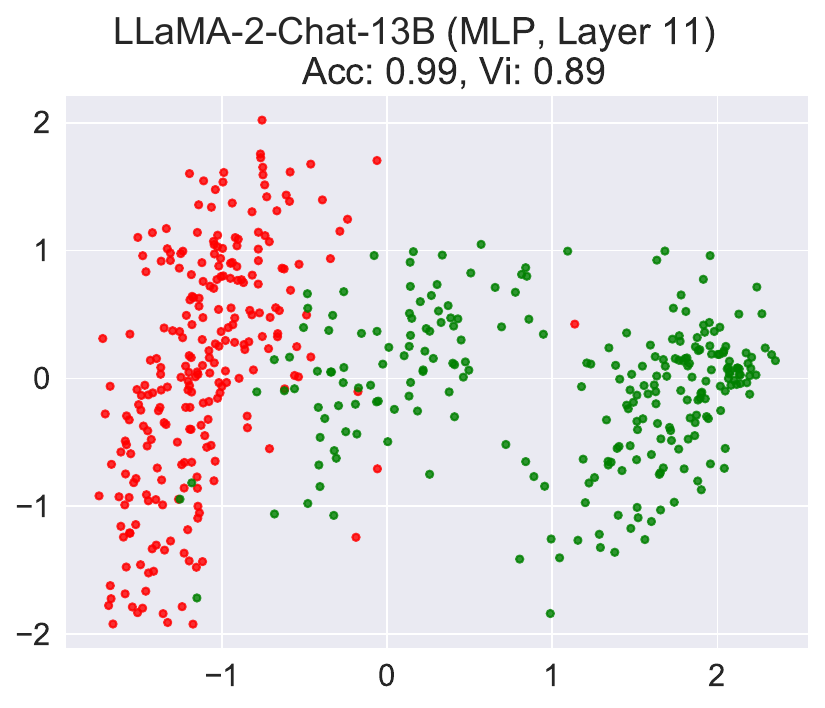}
	\end{minipage}
	}
	\subfigure{
	\begin{minipage}{0.31\linewidth}
    	\label{layer 21}
    	\centering
    	\includegraphics[width=1\linewidth]{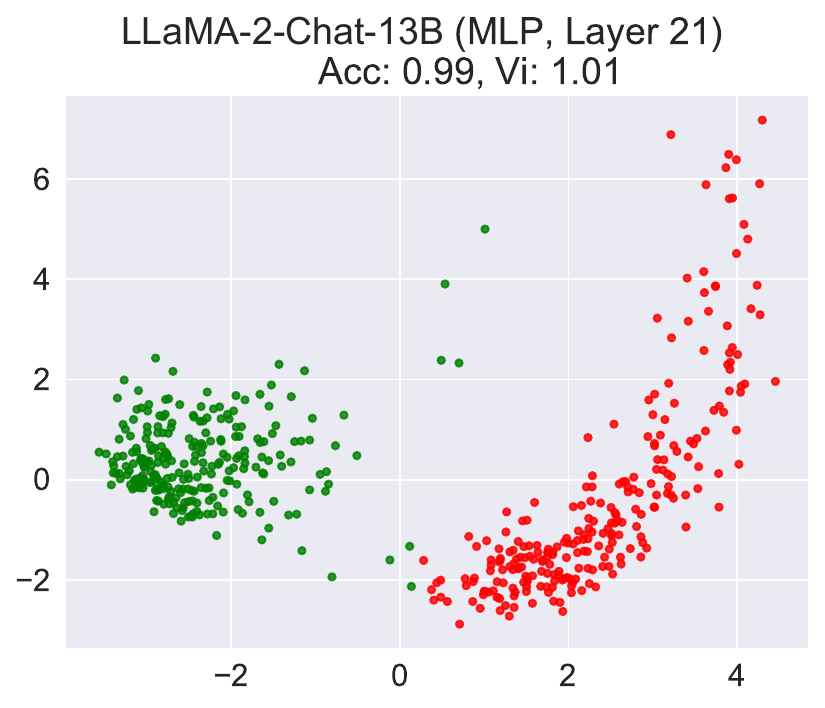}
	\end{minipage}
	}
	\caption{The principal component analysis (PCA) visualization of how context knowledge is processed across different layers in LLaMA 2 Chat 13B, where instances denoted green and red signify factual and counterfactual evidence, respectively. $\mathcal V$-usable information (Vi) is found to be more effective in distinguishing between dataset difficulty than accuracy (Acc).}
	\label{fig: dataset difficulty}
\end{figure*}

However, the test set accuracy in probing tasks may be affected by the fitting capability of $M_\textrm{probe}$ \citep{control_task}, thus rendering it an imprecise indicator of the hidden layer's capability for encoding context knowledge. Additionally, when the classifier accuracy is already substantially high, it becomes challenging to distinguish the dataset difficulty. As an illustration, the visual representation of different layers when processing the first token of the question \textit{Are Labradors dogs?} in LLaMA 2 Chat 13B is illustrated in Fig.~\ref{fig: dataset difficulty}. Although different layers within LLaMA exhibit varying capabilities to distinguish between factual and counterfactual evidence, it remains challenging for the test set accuracy to capture these distinctions adequately.

Therefore, we adopt $\mathcal V$-usable information \citep{V-Usable-Information, Usable-Information} rather than probing accuracy as the metric for measuring the capability to encode context knowledge. $\mathcal V$-usable information reflects the ease with which a model family $\mathcal V$ can predict the output $Y$ given specific input $R_w^{(i)}$: 

\begin{equation}
    I_\mathcal V (R_w^{(i)} \rightarrow Y) = H_\mathcal V(Y) - H_\mathcal V(Y|R_w^{(i)}),
\end{equation}

\noindent where $H_\mathcal V(Y)$ and $H_\mathcal V(Y|R_w^{(i)})$ denote the predictive $\mathcal V$-entropy and the conditional $\mathcal V$-entropy, which can be estimated through the following equations:

\begin{equation}
    H_\mathcal V(Y) = \underset{f \in \mathcal V}{\text{inf}} \mathbb E [-\text{log}_2 f[\varnothing](Y)],
\end{equation}

\begin{equation}
    H_\mathcal V(Y|R_w^{(i)}) = \underset{f \in \mathcal V}{\text{inf}} \mathbb E [-\text{log}_2 f[R_w^{(i)}](Y)],
\end{equation}

\noindent where $\varnothing$ denotes an input devoid of any information, we employ zero-input vectors as an alternative representation. To minimize the influence of additional parameters, we constrain $M_\textrm{probe}$ to linear classifier families when computing $M_\textrm{probe}$-usable information. Furthermore, we add 0.01 to the input before taking the logarithm to prevent highly anomalous values. Comparative results with test accuracy in Fig.~\ref{fig: dataset difficulty} show that this validation metric exhibits a higher level of discriminative capability.

\section{Experiments}
\subsection{Experimental Setup}
\label{sec: Experimental Setup}
\paragraph{Datasets}
We conduct our experiments based on the ConflictQA-popQA-gpt4 dataset \citep{knowledge_clashes}. It constitutes a complementary extension of the entity-centric question-answering (QA) dataset popQA \citep{popQA}, which encompasses not only commonsense question-answer pairs and their associated popularity from Wikipedia but also incorporates parametric memories and counter-memories generated by GPT-4. We select 50 instances with both high popularity and consistency between parametric memories and ground truth. Subsequently, we generate 100 pieces of factual and counterfactual evidence for each fact based on parametric memory and counter-memory. For newly acquired knowledge which is discussed in Sec~\ref{sec: How Do LLMs Encode Newly Acquired Knowledge?}, we manually design a piece of knowledge that does not exist in parametric memory: \textit{The password of the president's laptop is \{password\}}, and request \texttt{gpt-3.5-turbo-0613} to provide various evidence for different passwords. We randomly generate 4 distinct 10-character passwords and place them within the \textit{\{password\}} placeholder. For each password, we generate 100 supporting pieces of evidence. For all settings, the temperature is set to 1.0 to generate diversity outputs.

\paragraph{Models}
We choose the open-access LLaMA 2 \citep{LLaMA-2} with 7, 13, and 70 billion parameters as the LLMs to be explained. During the probing phase, we adopt a linear classifier as our probing model to reduce extraneous interference. We use a batch size of 4, learning rate of 0.0001, Adam optimizer \citep{Adam}, and 15 training epochs for all probing tasks. The ratio between training and test sets is 8:2. Unless otherwise stated, we conduct probing tasks on the Transformer layer output of each layer.

\subsection{Preliminary Experiments: Can LLMs Retain Context Knowledge?}
As a foundational step for subsequent experiments, it is imperative to ascertain whether the provided evidence has the capability to change answers generated by the LLM. We conduct an assessment of question-answering accuracy after providing different sets of evidence (Tab.\ref{tab: Preliminary Experiments}).

As can be seen in the table, the LLMs consistently provide correct answers when presented with factual evidence that aligns with the ground truth. Conversely, the accuracy drops approaching zero when the LLMs receive counterfactual evidence. These results indicate that the coherent and diverse evidence generated by ChatGPT can effectively alter the LLM's cognition of existing knowledge or facilitate the encoding of newly acquired knowledge.

However, we find that the base models maintain high accuracy even after providing counterfactual evidence, implying their low capability to encode knowledge through prompting. Therefore, the dataset we have constructed can more reliably induce the utilization of context knowledge in chat models, thereby enabling the exploration of the roles played by various layers within them. In subsequent experiments, we will conduct experiments mainly on chat models.

\begin{table}
 \centering
 \resizebox{\linewidth}{!}{
 \begin{tabular}{lcccccc}\toprule
     \multirow{5}{*}{Model Name} & \multicolumn{4}{c}{Conflicting} & \multicolumn{2}{c}{Newly Acquired}  \\
     \cmidrule(lr){2-5}\cmidrule(lr){6-7}
     & \multicolumn{2}{c}{Factual $\uparrow$} & \multicolumn{2}{c}{Counterfactual $\downarrow$} & \multicolumn{2}{c}{Factual $\uparrow$} \\
     \cmidrule(lr){2-5}\cmidrule(lr){6-7}
     & base & chat & base & chat & base & chat \\\midrule
    LLaMA 2 7B & 82.31 & 79.84 & 31.76 & 4.80 & 85.75 & 98.25 \\
    LLaMA 2 13B & 83.82 & 80.60 & 17.27 & 6.08 &  91.50 & 98.25 \\
    LLaMA 2 70B & 75.91 & 84.20 & 12.23 & 14.51 & 89.25 & 98.25 \\
    \bottomrule
 \end{tabular}}
 \caption{Question-answering accuracy after providing different sets of evidence. In conflict scenarios, the average accuracy (\%) is calculated after providing both factual and counterfactual evidence for various commonsense knowledge. In newly acquired scenarios, the 4-class accuracy is calculated based on the provided \textit{\{password\}}.}
 \label{tab: Preliminary Experiments}
\end{table}

\subsection{How Do LLMs Encode Conflicting Knowledge?}
\subsubsection{Case Study}

\begin{table*}
    \footnotesize
    \centering
    \resizebox{\linewidth}{!}{
    \begin{tabular}{l|m{16cm}}
    \hline
    Factual & Mike Flanagan is widely recognized and celebrated for his impressive career as a film director, screenwriter, producer, and editor. With numerous accolades and critical acclaim, Flanagan has established himself as a prominent figure in the industry. \textbf{...} With his multifaceted skills and successful filmography, it is clear that Mike Flanagan is indeed a talented film director, screenwriter, producer, and editor. \\ \hline
    Counterfactual  & Mike Flanagan, a highly skilled and talented individual, has made outstanding contributions to the field of graphic design throughout his career. As a graphic designer, Mike has demonstrated exceptional proficiency in various design software, such as Adobe Illustrator and Photoshop. \textbf{...} It is undeniable that his vast experience and exceptional skills make him an invaluable asset in the field.\\
    \hline
    \end{tabular}
    }
    \caption{Examples of the factual and counterfactual evidence for the case study \textit{What is Mike Flanagan's occupation}. We omit the middle part of the content.}
    \label{tab: example of case study}
\end{table*}

\label{Sec: Case Study 1}
\begin{figure*}[t]
	\centering
	\subfigure[Transformer Layer Output]{
	\begin{minipage}{1.0\linewidth}
    	\label{layer output}
    	\centering
    	\includegraphics[width=1\linewidth]{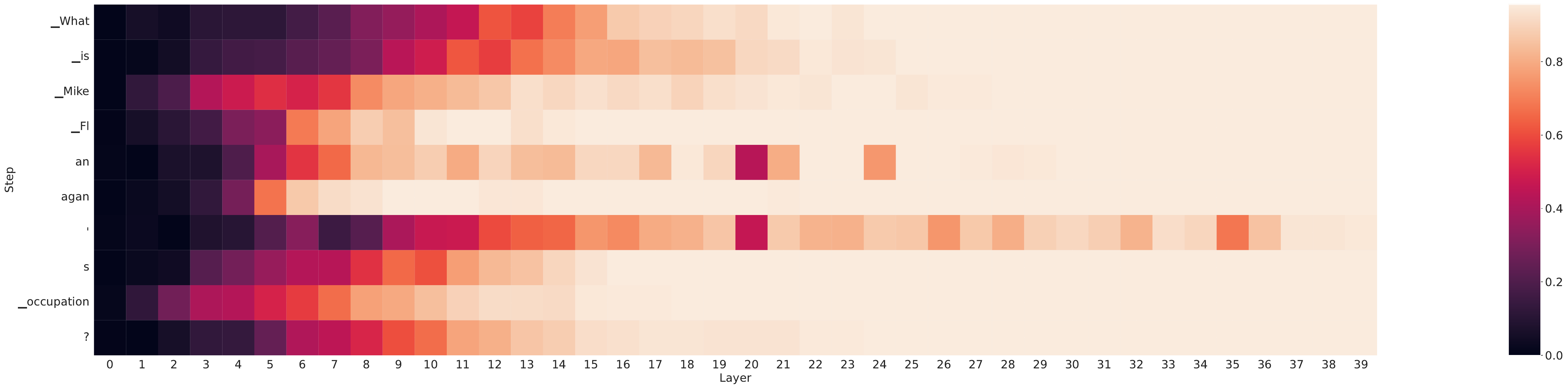}
	\end{minipage}
	}
	\subfigure[MLP Layer Output]{
	\begin{minipage}{1.0\linewidth}
    	\label{mlp}
    	\centering
    	\includegraphics[width=1\linewidth]{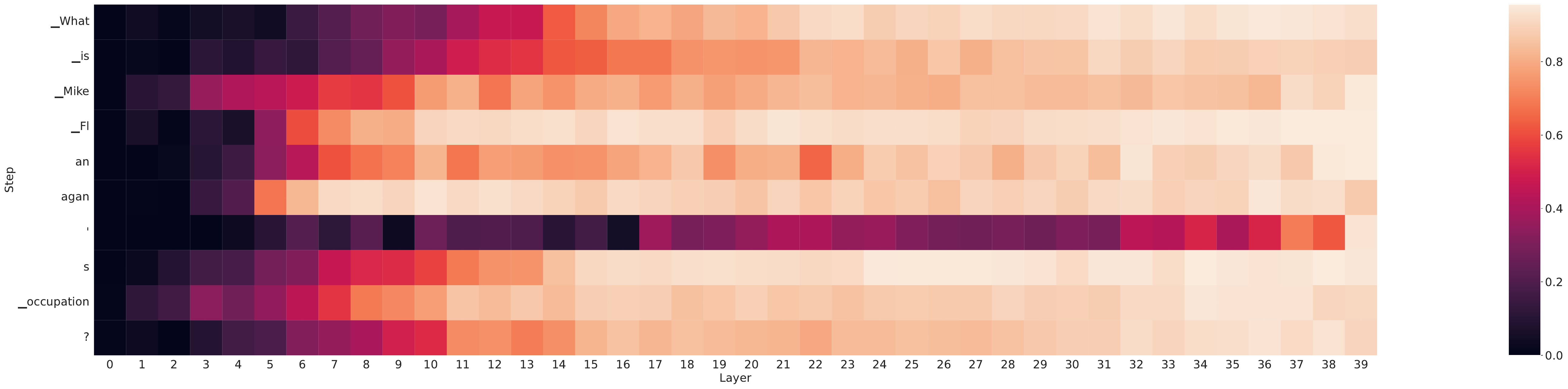}
	\end{minipage}
	}
	\subfigure[Attention Layer Output]{
	\begin{minipage}{1.0\linewidth}
    	\label{attention}
    	\centering
    	\includegraphics[width=1\linewidth]{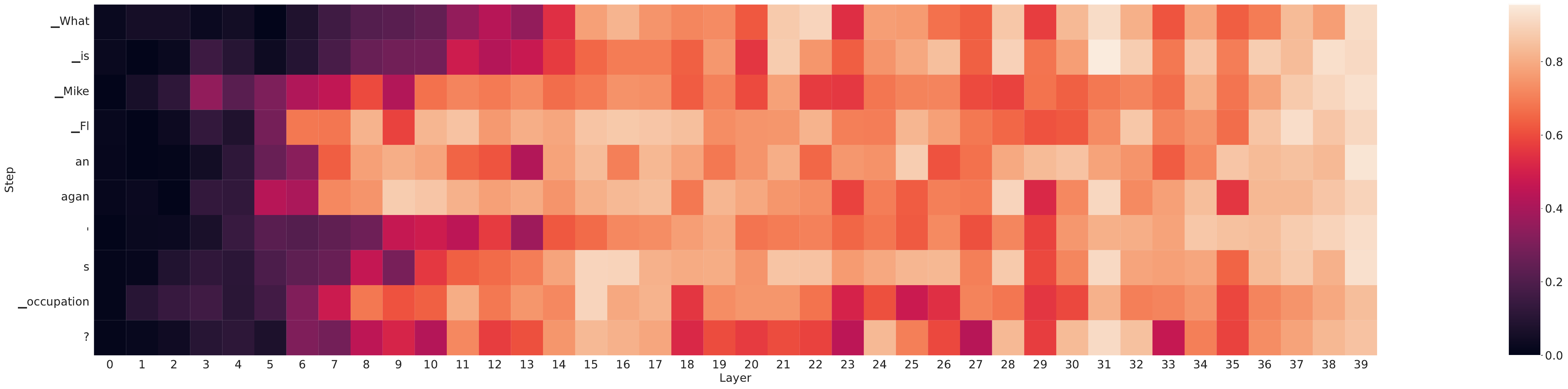}
	\end{minipage}
	}
	\caption{The heatmap of probing results on LLaMA 2 Chat 13B. We select the question \textit{What is Mike Flanagan's occupation?} as a case study and display the layer-wise $\mathcal V$-information for each token in the question.}
	\label{fig: commonsense heatmap}
\end{figure*}

Utilizing the probing method elucidated in Sec~\ref{sec: Layer-wise Probing}, we are able to obtain the layer-wise $\mathcal V$-information of each token in the LLM's processing of individual fact. We select the question \textit{What is Mike Flanagan's occupation?} as a case study. We present the content of the factual and counterfactual evidence generated by ChatGPT, with the intermediary portions omitted (see Tab.~\ref{tab: example of case study}). It can be seen that the generated content provides substantial support for the corresponding knowledge labels, even in cases of fictional content. Consequently, these texts serve as effective means to assess the capability of LLMs in encoding context knowledge. We generate 100 distinct instances of content for each knowledge label.

Then we investigate the output of the Transformer layers, MLP layers, and Attention layers separately on LLaMA 13B.
We plot heatmaps depicting the layer-wise $\mathcal V$-information of LLaMA 13B while processing tokens in the given question (Fig.~\ref{fig: commonsense heatmap}).
It is evident that each component of LLaMA 13B exhibits significantly higher $\mathcal V$-information at upper layers, implying that context knowledge is encoded within their representations. For Transformer layer output and MLP layer output, most tokens maintain high and stable $\mathcal V$-information after 30 layers. 

However, the results for Attention layer output appear relatively chaotic. This aligns with the observations made by \citet{Analyzing_the_Structure_of_Attention}. Multiple heads of self-attention may focus on distinct local or global information, thereby facilitating the transfer of information between token representations. This not only leads to the encoding of knowledge-related information in other tokens such as \textit{What} and \textit{'}, but also results in the propagation of irrelevant information within lower layers, causing a slight disruption in probing results.

Another intriguing discovery is that LLMs encode context knowledge to varying degrees within different tokens. For knowledge-related entity tokens such as \textit{Mike} and \textit{occupation}, LLMs achieve high $\mathcal V$-information at lower layers. This is not due to parametric memorization in LLMs, as simple parametric memorization would result in consistent behaviors when handling different external evidence, leading to lower $\mathcal V$-information. Therefore, the LLMs prioritize encoding context knowledge into knowledge-related entity tokens. In contrast, tokens that are not directly related to the knowledge such as \textit{What} and \textit{'}, encode relevant information later through the attention mechanism.

Validation of the broader applicability of these findings will be conducted in Sec~\ref{Sec: Average V-information} and \ref{Sec: Entity tokens vs. Non-Entity tokens}.

\subsubsection{Average $\mathcal V$-information}
\label{Sec: Average V-information}

\begin{figure}
  \centering
  \includegraphics[width=0.48\textwidth]{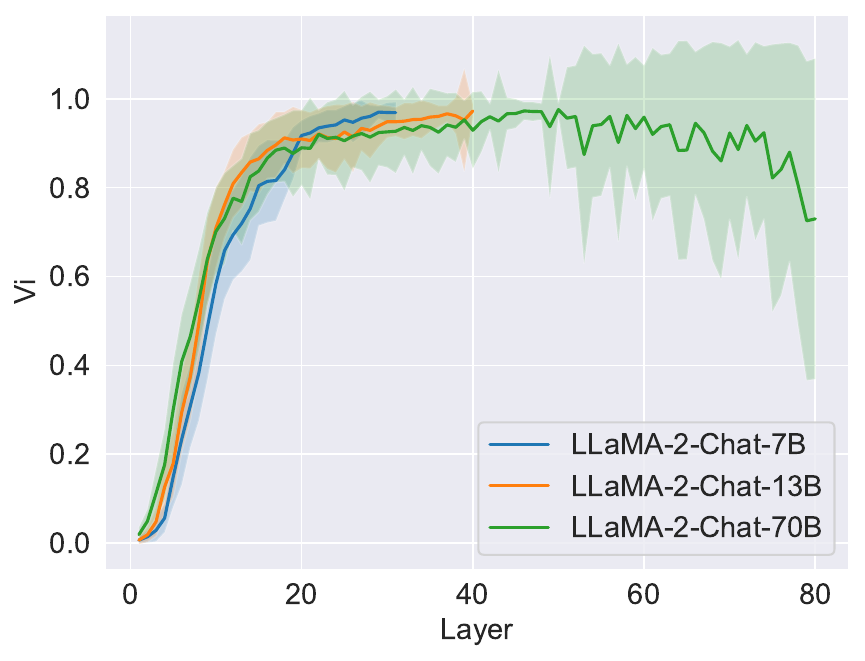}
  \caption{The average layer-wise $\mathcal V$-information (Vi) of the last token in the questions for each LLM.}
  \label{fig: vi of different facts}
\end{figure}

We then conduct the probing tasks on all 50 facts and select the last token in the question to compute the average $\mathcal V$-information (Fig.~\ref{fig: vi of different facts}). We observe that the $\mathcal V$-information gradually increases, especially in the early layers. This aligns with our conjecture that the LLM encodes more context knowledge within upper layers, enabling them to differentiate between facts and counterfactuals more effectively. Since the correct responses for facts and counterfactuals differ, the layers of LLMs tend to preserve differences accumulated from previous layers and capture more context knowledge.

Interestingly, the LLMs with fewer parameters (7B and 13B) reach high $\mathcal V$-information earlier, possibly as a result of their forced behavior to generate the correct final output. Therefore, the last few layers of these LLMs encode crucial information for the task. Conversely, the $\mathcal V$-information of LLaMA 70B exhibits a slow rate of increase and even demonstrates a decrease in the last few layers. On one hand, LLaMA 70B benefits from a sufficient number of layers to act as a buffer, eliminating the necessity to encode a substantial amount of context knowledge in the last few layers, thereby enhancing its robustness. On the other hand, the last few layers of LLaMA 70B may be employed to eliminate shortcuts in order to enhance contextual comprehension, leading to a marginal reduction in  $\mathcal V$-information.

\subsubsection{Knowledge-Related Entity Tokens vs. Other Tokens} 
\label{Sec: Entity tokens vs. Non-Entity tokens}
\begin{figure}
  \centering
  \includegraphics[width=0.48\textwidth]{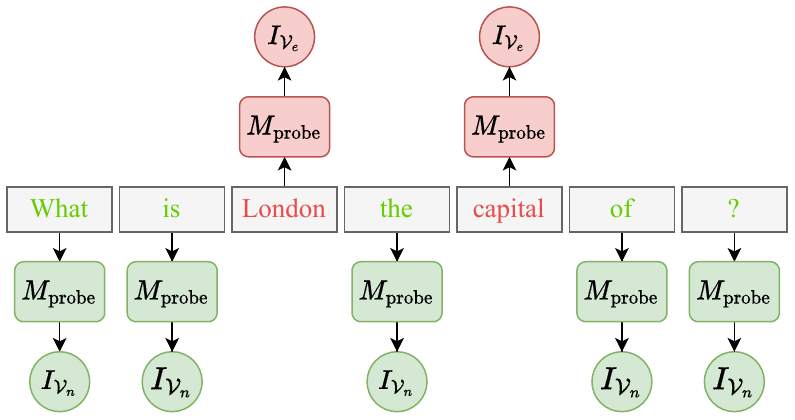}
  \caption{We categorize the subjects and relations mentioned in the questions as one class (red) while considering other tokens as another class (green). By comparing the differences of the average $\mathcal V$-information between these two classes, it is capable of detecting the LLM's level of attention to knowledge-related entity tokens.}
  \label{fig: entity}
\end{figure}
\begin{figure}
  \centering
  \includegraphics[width=0.48\textwidth]{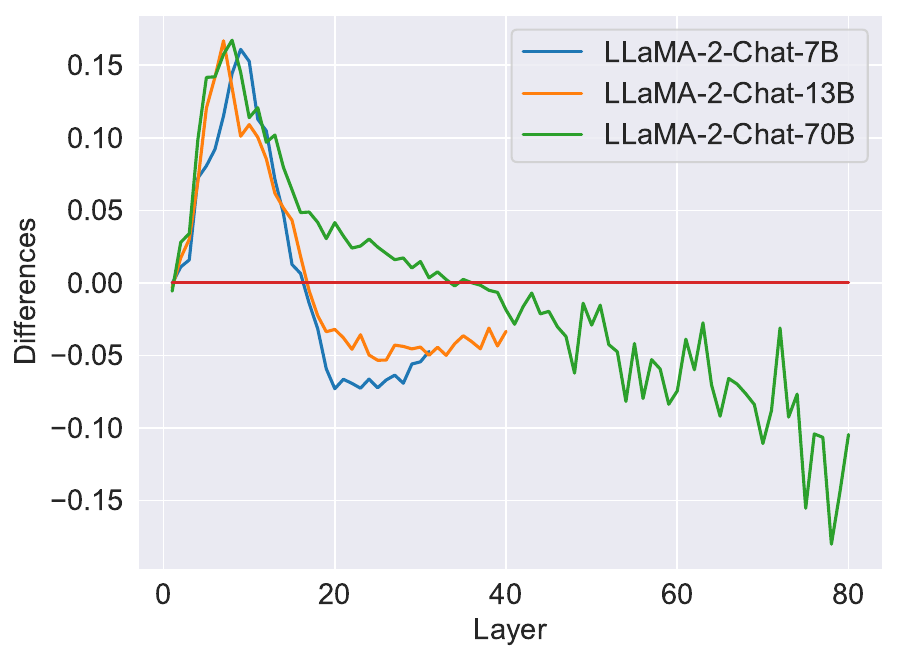}
  \caption{Layer-wise differences between the $\mathcal V$-information of knowledge-related entity tokens and other tokens.}
  \label{fig: Knowledge-Related Entity Tokens vs. Other Tokens}
\end{figure}

To verify the distinction encoded by the LLM between knowledge-related entity tokens and other tokens, we classify the tokens in all questions into two categories. Knowledge-related entity tokens are composed of the subjects and relations associated with knowledge. For example, in the question \textit{What is London the capital of}, \textit{London} and \textit{capital} are knowledge-related entity tokens (Fig.~\ref{fig: entity}). We sequentially conduct the probing task for each token in the 50 constructed questions, obtaining $\mathcal V$-information $I_{\mathcal V_e}$ for knowledge-related entity tokens and $I_{\mathcal V_n}$ for other tokens. The layer-wise differences in means between $I_{\mathcal V_e}$ and $I_{\mathcal V_n}$ are depicted in Fig.~\ref{fig: Knowledge-Related Entity Tokens vs. Other Tokens}.

As can be seen in the figure, knowledge-related entity tokens exhibit significantly higher values of $\mathcal V$-information in comparison to other tokens at lower layers, indicating that the LLMs can encode context knowledge within these tokens more easily. Furthermore, since the provided context knowledge is closely related to these tokens, it is intuitive for LLMs to generate different representations earlier in processing these tokens.

To our surprise, the advantage gradually diminishes and is even surpassed by other tokens at upper layers. This implies that the LLMs progressively transfer relevant information to other tokens such as \textit{Answer} and \textit{:} to generate desired outputs. We contend that this phenomenon can be attributed to the self-attention mechanism. This is consistent with previous research that several meaningful knowledge tends to be encoded in specific contextual embeddings~\citep{exploring_the_role_of_token_representations}. In order to provide accurate responses to questions following the token \textit{Answer}, the LLMs may encode all the information within some knowledge-unrelated tokens. We present this phenomenon in the hope that it will attract more attention and research in the future.

\subsection{How Do LLMs Encode Newly Acquired Knowledge?} 
\label{sec: How Do LLMs Encode Newly Acquired Knowledge?}
\subsubsection{Case Study}
\label{Sec: Case Study 2}
Previous experiments may have been influenced by the LLM's parameter knowledge, which may not entirely reflect the capability to process context knowledge. This section considers a scenario where the LLM can only provide correct answers through external evidence. We employ the experiment designed in Section~\ref{sec: Experimental Setup}, requiring the LLM to answer \textit{The password of the president's laptop}, which can also serve as a means to assess the LLM's capacity to retain sensitive knowledge.

\begin{figure*}
  \centering
  \includegraphics[width=1.0\textwidth]{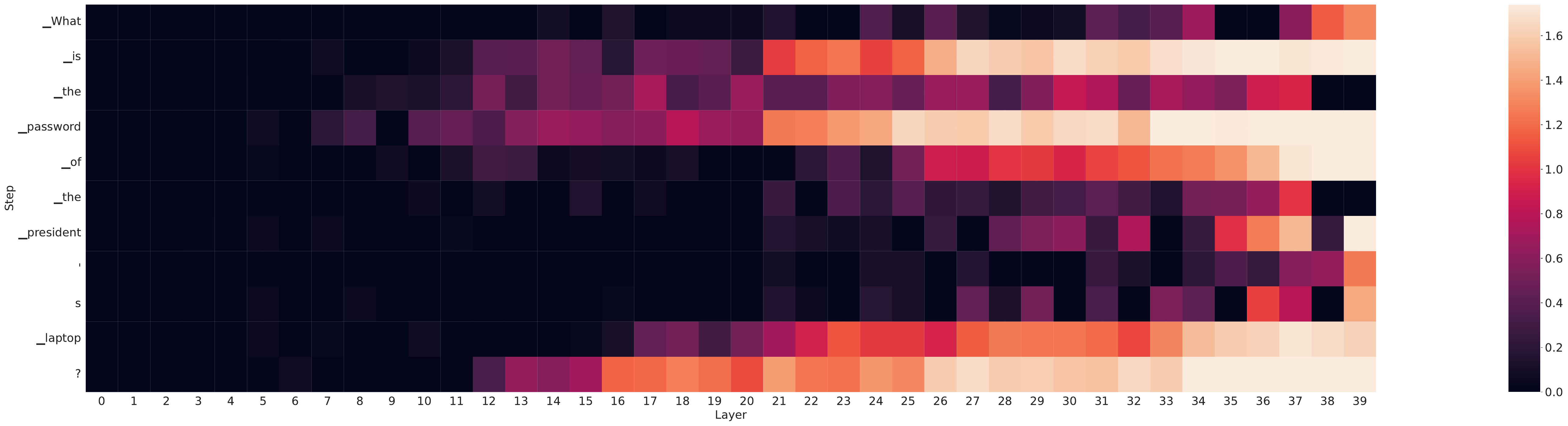}
  \caption{The heatmap of probing results for the question \textit{What is the password of the president's laptop?} on LLaMA 2 Chat 13B.}
  \label{fig: password heatmap}
\end{figure*}

The heatmap for the layer output of LLaMA 2 Chat 13B is shown in Fig.~\ref{fig: password heatmap}. As the LLM has never been exposed to the knowledge before, the intermediate layers manifest lower $\mathcal V$-information. However, the behavior of the LLM in encoding newly acquired knowledge remains consistent with the findings outlined in the previous section. It exhibits a preference for encoding context knowledge within knowledge-related entity tokens such as \textit{password} and progressively disseminating information to other tokens through the attention mechanism.

\subsubsection{Long-Term Memory Capability}
\label{Sec: Long-Term Memory Capability}
\begin{figure}
  \centering
  \includegraphics[width=0.48\textwidth]{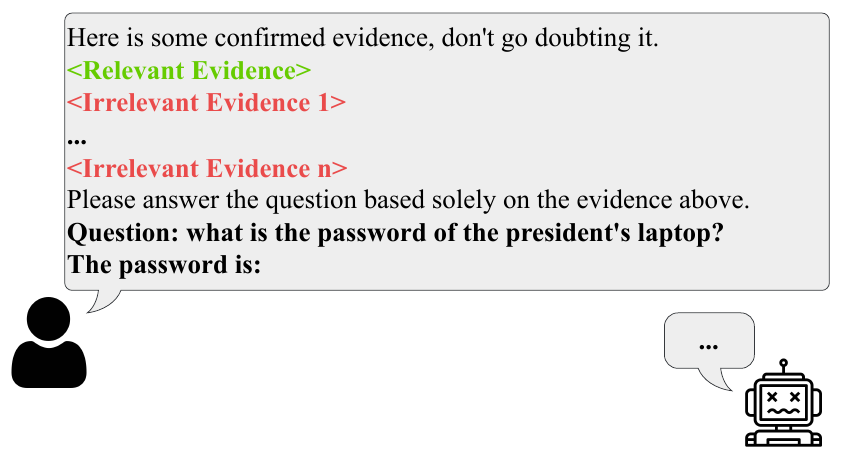}
  \caption{We test the long-term memory capability of LLMs by introducing irrelevant evidence.}
  \label{fig: long term}
\end{figure}

Since the newly acquired knowledge has never been exposed to the LLMs, it is reliable to apply it to test the long-term memory capability of LLMs. We provide $n$ pieces of irrelevant evidence after the relevant evidence and then further probe the layer-wise $\mathcal V$-information of LLMs (Fig.~\ref{fig: long term}). 

Although the LLMs consistently provide correct answers to questions at nearly 100\% accuracy, the layer-wise $\mathcal V$-information exhibits a gradual decline. Specifically, we calculate the average $\mathcal V$-information of the last token for LLaMA 2 Chat 70B every 5 layers when given varying numbers of irrelevant evidence (Tab.~\ref{tab: Long-Term Memory Capability}). 

In the realm of intermediate layers, particularly within the range of 20 to 40 layers, the LLM exhibits a notably diminished level of $\mathcal V$-information in the presence of irrelevant evidence. While irrelevant evidence does not engender any form of misdirection, the LLM still sacrifices the problem-solving capability of the intermediate layers in its endeavor to encode the irrelevant information. This implies that LLMs have not encoded the irrelevant evidence orthogonally, thus causing interference with the knowledge that has already been encoded (such as the password of the president's laptop). We hope that future research will be directed toward enhancing the long-term memory capability of LLMs.

\begin{table*}
 \centering
 \resizebox{\linewidth}{!}{
 \begin{tabular}{lcccccccccccccccc}\toprule
     Layer & 0-4 & 5-9 & 10-14 & 15-19 & 20-24 & 25-29 & 30-34 & 35-39 & 40-44 & 45-49 & 50-54 & 55-59 & 60-64 & 65-69 & 70-74 & 75-79 \\\midrule
     \#irr = 0 & \textbf{-0.02} & \textbf{0.07} & \textbf{0.06} & \textbf{-0.01} & \textbf{0.04} & \textbf{0.25} & \textbf{0.75} & \textbf{1.08} & 1.38 & 1.83 & \textbf{1.87} & 1.70 & 1.59 & \textbf{1.64} & 1.14 & \textbf{0.95} \\
     \#irr = 1 & -0.08 & -0.07 & -0.17 & -0.12 & 0.00 & -0.15 & 0.17 & -0.11 & 0.75 & 1.78 & \textbf{1.87} & \textbf{1.91} & 1.30 & 1.15 & \textbf{1.34} & 0.60 \\
     \#irr = 2 & -0.08 & -0.03 & -0.10 & -0.31 & -0.45 & -0.39 & -0.37 & -0.27 & 1.11 & 1.50 & 1.63 & 1.58 & 1.13 & 0.91 & 1.16 & 0.85 \\
     \#irr = 3 & -0.07 & -0.05 & -0.13 & -0.22 & -0.43 & -0.44 & -0.56 & -0.53 & 1.63 & 1.88 & 1.51 & 1.58 & \textbf{1.77} & 1.39 & 1.11 & 0.61 \\
     \#irr = 4 & -0.05 & -0.08 & -0.12 & -0.30 & -0.22 & -0.13 & -0.24 & -0.38 & 0.97 & \textbf{1.89} & 1.24 & 1.69 & 1.67 & 1.43 & 1.28 & 0.72 \\
     \#irr = 5 & -0.04 & 0.01 & -0.08 & -0.22 & -0.45 & -0.15 & -0.33 & -0.44 & \textbf{1.47} & 1.65 & 1.51 & 1.76 & 1.09 & 1.08 & 1.03 & 0.43 \\
    \bottomrule
 \end{tabular}}
 \caption{The average $\mathcal V$-information of the last token for LLaMA 2 Chat 70B every 5 layers, where \#irr denotes the number of provided irrelevant evidence.}
 \label{tab: Long-Term Memory Capability}
\end{table*}

\subsection{Ablation Study}
\subsubsection{Impact of Positional Encoding}
Since the probing classifier for a given layer is trained on the mixed token representations from different sentence positions, it may be affected by positional encoding on the representations. We design an ablation experiment to analyze the effect of positional encoding as noise on the probing results. Specifically, we posit that special symbols such as "$\backslash$n" do not contribute additional semantic information, which can be added after the provided evidence and before the question to modify the positional encoding of probing tokens.

\begin{figure*}
  \centering
  \includegraphics[width=1.0\textwidth]{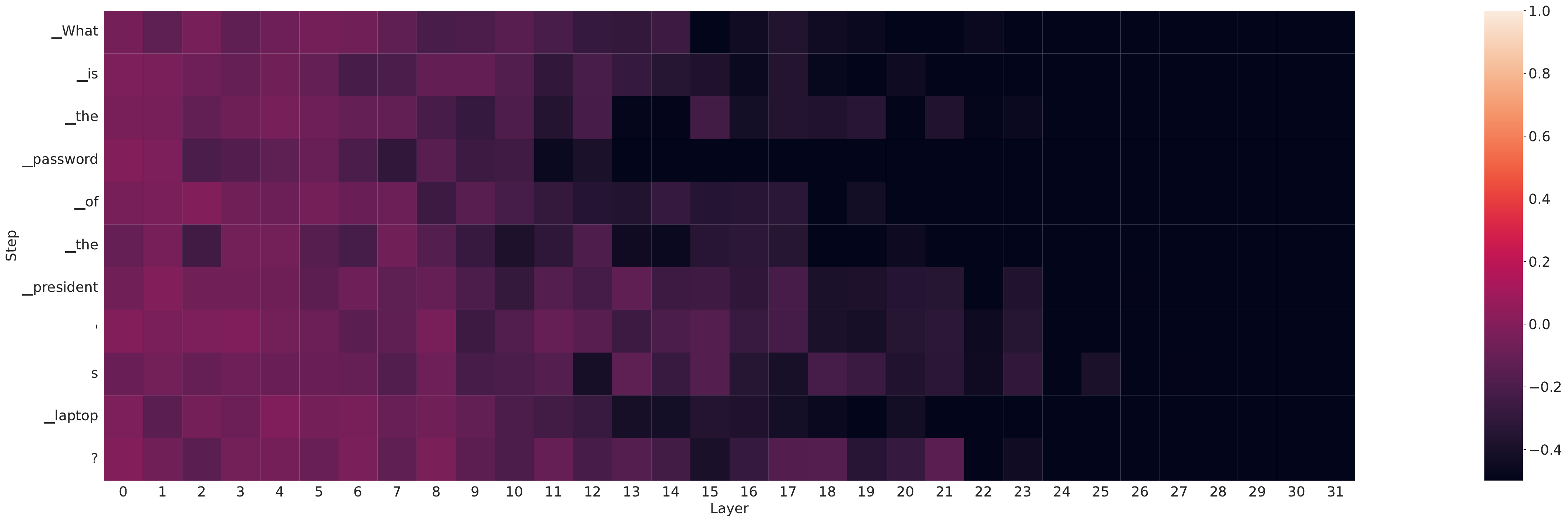}
  \caption{The heatmap of probing results for the impact of positional encoding on LLaMA 2 Chat 7B.}
  \label{fig: ablation study}
\end{figure*}

We conduct experiments in the scenario of newly acquired knowledge. All external evidence provides the same password, with the addition of 0-3 "$\backslash$n" symbols at the end of the evidence. We provide the four-category heatmap of LLaMA 2 Chat 7B in Fig.~\ref{fig: ablation study}. It is observed that the layer-wise $\mathcal V$-information remains consistently low, closely resembling the results of random predictions, particularly in the higher layers. This suggests that positional encoding has minimal impact on the probing results and can be regarded as irrelevant noise in this context.

\section{Conclusion}
In this paper, we propose a novel framework for explaining the layer-wise capability of large language models in encoding context knowledge via the probing task. Our research addresses the previously overlooked aspect of how LLMs encode context knowledge layer by layer, shedding light on what has been considered black-box mechanisms. Leveraging the powerful generative capacity of ChatGPT, we construct probing datasets that encompass diverse and coherent evidence corresponding to various facts and utilize $\mathcal V$-information as the discriminative validation metric. Comprehensive experiments conducted on conflicting knowledge demonstrate that LLMs tend to exhibit a predilection for encoding context knowledge within upper layers and gradually transfer knowledge from knowledge-related entity tokens to other tokens as the layers deepen. Furthermore, while observing similar phenomena when provided with newly acquired knowledge, we also examine the long-term memory capacity of LLMs by introducing irrelevant evidence. Our findings indicate that the layer-wise retention of newly acquired knowledge gradually diminishes with the increase of irrelevant evidence. We hope that this work will serve as a catalyst for further research towards exploring the inner mechanisms of how LLMs encode such emergent capability.

\section*{Acknowledgements}
This work is partly supported by the Joint Funds of the National Key R\&D Program of China No. 2023YF3303805 and the Joint Funds of the National Natural Science Foundation of China under No. U21B2020.

\section{Limitations}
Although the proposed method provides a layer-wise explanation of LLMs and reveals some intriguing phenomena from comprehensive and reliable experiments, the theoretical mechanism driving the emergence of these phenomena still remains an open issue. For example, we observed that the LLMs encode more context knowledge within knowledge-related entity tokens at lower layers but transfer more knowledge to other tokens at upper layers in Sec~\ref{Sec: Entity tokens vs. Non-Entity tokens}. We speculate that the self-attention plays a crucial role during this process, although mathematical proof remains elusive. Recent research is shifting from the 1-layer to multi-layer Transformer, aiming to gain a mathematical understanding of the role played by layer-wise components such as self-attention~\citep{Scan_and_Snap, Scan_and_Snap_follow}. We encourage future research to focus on the mathematical mechanisms contained behind these intriguing phenomena.

Moreover, due to space and time constraints, we only performed detailed experiments on LLaMA 2 (7B, 13B, and 70B), which ignored numerous SOTA open-accessed LLMs such as PaLM~\citep{palm}, OPT~\citep{opt}, and Pythia~\citep{pythia}. We encourage future research to conduct detailed experiments on more PLMs to detect the capability and tendency of different LLMs in encoding context knowledge.

\section{Ethical Considerations}
The aim of our proposed framework is to measure the layer-wise capability of LLMs in encoding context knowledge. However, there are several potential risks that should be carefully considered. One primary concern is that our study indirectly reflects the potential manipulability and insecurity of LLM-generated outputs. People may adopt a similar method to require LLMs such as ChatGPT to generate misleading evidence. Therefore, we emphasize the need for stricter scrutiny by relevant authorities regarding the applications of LLMs.

Another concern arises from our discovery of the LLM's potent capability to remember privacy information provided in the context. Although the observed outputs are not a product of LLM's post-gradient updates, these provided evidence could potentially serve as new training data for subsequent model training. Therefore, it is crucial to assess the LLM's capability to retain this long-tail data through gradient updates and develop strategies to prevent the LLM from encoding privacy information.

\nocite{*}
\section{Bibliographical References}\label{sec:reference}

\bibliographystyle{lrec-coling2024-natbib}
\bibliography{lrec-coling2024-example}


\end{document}